\documentclass[sigconf]{acmart}
\AtBeginDocument{%
  }

\setcopyright{acmlicensed}
\copyrightyear{2018}
\acmYear{2018}
\acmDOI{XXXXXXX.XXXXXXX}
\acmConference[Conference acronym 'XX]{Make sure to enter the correct
  conference title from your rights confirmation email}{June 03--05,
  2018}{Woodstock, NY}
\acmISBN{978-1-4503-XXXX-X/2018/06}

\usepackage{subcaption}

\begin{document}

\title{Learning to Erase Private Knowledge from Multi-Documents for Retrieval-Augmented Large Language Models}

\author{Yujing Wang}
\authornote{Both authors contributed equally to this research.}
\email{Eugenia@buaa.edu.cn}
\orcid{1234-5678-9012}
\author{Jinwen Chen}
\authornotemark[1]
\email{jwkami@buaa.edu.cn}
\affiliation{%
  \institution{Beijing Advanced Innovation Center for Future Blockchain and Privacy Computing}
  \institution{School of Artificial Intelligence, Beihang University, China}
  \city{Beijing}
  \country{China}
}

\author{Hainan Zhang}
\affiliation{%
  \institution{Beijing Advanced Innovation Center for Future Blockchain and Privacy Computing}
  \institution{School of Artificial Intelligence, Beihang University, China}
  \city{Beijing}
  \country{China}
\email{zhanghainan1990@163.com}
}

\author{Liang Pang}
\affiliation{%
  \institution{Institute of Computing Technology, Chinese Academy of Sciences}
  \city{Beijing}
  \country{China}
}

\author{Yongxin Tong}
\affiliation{%
 \institution{School of Computer Science and Engineering, Beihang University}
  \city{Beijing}
  \country{China}
}

\author{Binghui Guo}
\affiliation{%
  \institution{Institute of Artificial Intelligence, Beihang University}
  \city{Beijing}
  \country{China}
  }

\author{Hongwei Zheng}
\affiliation{%
  \institution{Beijing Academy of Blockchain and Edge Computing}
  \city{Beijing}
  \country{China}
}

\author{Zhiming Zheng}
\affiliation{%
  \institution{Beijing Advanced Innovation Center for Future Blockchain and Privacy Computing}
  \institution{Institute of Artificial Intelligence, Beihang University}
  \city{Beijing}
  \country{China}
}

\renewcommand{\shortauthors}{Trovato et al.}

\begin{abstract}
Retrieval-Augmented Generation (RAG) is a promising technique for applying LLMs to proprietary domains. However, retrieved documents may contain sensitive knowledge, posing risks of privacy leakage in generative results.
Thus, effectively erasing private information from retrieved documents is a key challenge for RAG. 
Unlike traditional text anonymization, RAG should consider: (1) the inherent multi-document reasoning may face de-anonymization attacks; (2) private knowledge varies by scenarios, so users should be allowed to customize which information to erase; (3) preserving sufficient publicly available knowledge for generation tasks.
This paper introduces the privacy erasure task for RAG and proposes Eraser4RAG, a private knowledge eraser that effectively removes user-defined private knowledge from documents while preserving sufficient public knowledge for generation. Specifically, we first construct a global knowledge graph to identify potential knowledge across documents, aiming to defend against de-anonymization attacks. Then we randomly split it into private and public sub-graphs, and fine-tune Flan-T5 to rewrite the retrieved documents, excluding private triples. Finally, PPO algorithm optimizes the rewriting model to minimize private triples and maximize public triples retention.
Experiments on four QA datasets demonstrate that Eraser4RAG achieves a strong privacy--utility balance among document-level rewriting methods, outperforming GPT-4o-based rewriting in private knowledge erasure while preserving substantially more public knowledge than aggressive keyword- or entity-suppression baselines.
\end{abstract}



\received{20 February 2007}
\received[revised]{12 March 2009}
\received[accepted]{5 June 2009}

\maketitle

\section{Introduction}
Training Large Language Models(LLMs) requires substantial computational resources and centralized data storage, which limits their application in proprietary and private domains~\cite{ye2024openfedllm,kuang2024federatedscope,zheng2024safely}. 
Retrieval-augmented generation (RAG) is regarded as a promising solution to deploy large language models (LLMs) in proprietary domains~\cite{fan2024survey,rw7,rw15}, as it leverages proprietary data as a retrieval corpus to obtain relevant documents for generation.

\begin{figure}[!t]
    \centering
    \includegraphics[width=\linewidth]{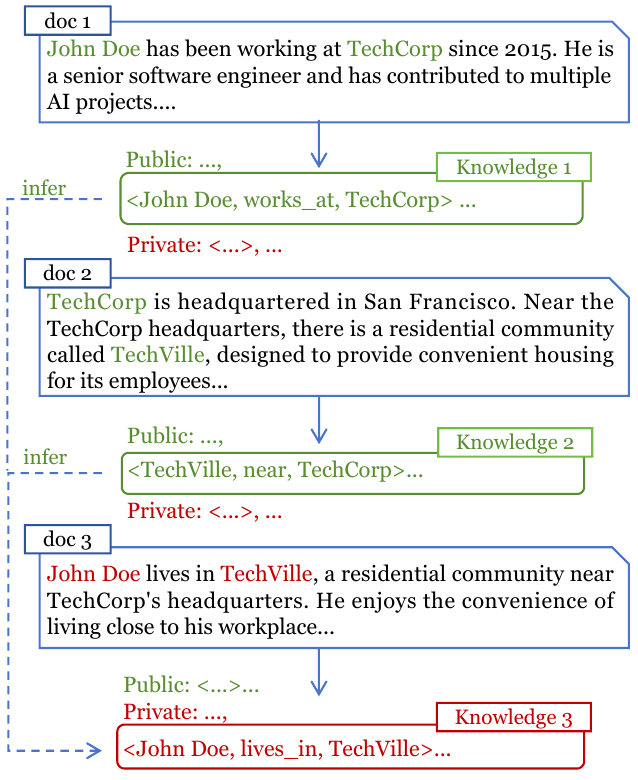}
    \caption{Example of de-anonymization through multiple documents reasoning.}
    \Description{Example of de-anonymization through multiple documents reasoning.}
    \label{fig:multi_doc}
\end{figure}

However, documents retrieved from proprietary data may contain sensitive information, such as home addresses or social relationships, posing risks of privacy leakage in generative results~\cite{rw7}. Therefore, it is crucial to design an efficient private knowledge eraser for RAG, capable of removing sensitive information from retrieval documents while preserving public knowledge to support downstream generation. Traditional text anonymization methods~\cite{dou-etal-2024-reducing, frikha2024incognitext, mosallanezhad2019deep} rely on obfuscating expressions or adding noise to erase specific private knowledge in a document. They leverage differential privacy techniques that add noise during training and tokenization to remove private information, or use synthetic data generation, document sanitization, and privacy-aware retrieval to preserve utility while minimizing exposure of private information. 

Unlike traditional text anonymization tasks, the \textbf{Privacy Erasure} task in RAG scenarios presents more complex challenges: (1) The combination of multiple documents can pose a risk of de-anonymization attacks~\cite{ji2016graph}. Even if a single document does not contain private information, aggregating multiple documents can still lead to the inference of private knowledge~\cite{hoang2023protecting}. As shown in Figure~\ref{fig:multi_doc}, doc1 and doc2 each contain public knowledge triples, Knowledge1 and Knowledge2. But these public triples together can be used to infer Knowledge3, which involves private information about "the living address of John Doe". (2) The definition of private knowledge varies across different scenarios, e.g., a celebrity's birthday is public, but a teenager's is not. Privacy information not only includes standalone non-relational private entities (such as Personally Identifiable Information, PII), but also fine-grained relational knowledge across the retrieved documents. Therefore, a human-centric RAG eraser should allow precise, controllable removal based on user definition. 
(3) Retaining adequate public knowledge is crucial for generative model reasoning, to ensure RAG remains practical and valuable.

In this paper, we propose \textbf{Eraser4RAG}, a private knowledge Eraser that rewrites documents to maximize public knowledge while minimizing user-defined private knowledge. Specifically, we first extract knowledge triples and entities from each retrieved document and construct a global knowledge graph to identify potential private knowledge across documents, reducing de-anonymization risks. Then, a connected subset of the graph is selected as the private graph, and the remainder forms the public graph. Given the original documents, private graph, and public graph, we fine-tune Flan-T5 model to generate rewritten documents excluding private triples and entities. Finally, we explore the optimal rewriting strategy for Eraser4RAG using PPO algorithm to maximize the retention of public knowledge and minimize the presence of private knowledge. During inference, the private knowledge triples or entities are firstly provided by the upstream data controller according to domain- or user-specific requirement. And then, our rewriting model focuses on removing the annotated private content from the retrieved documents while retaining valuable public knowledge for downstream QA tasks.

Experimental results on four public QA datasets demonstrate that Eraser4RAG achieves better erase performance than GPT-4o-based rewriting, with the public triple ratio increasing by 1.8\% and the private triple ratio decreasing by 21.8\%. Compared with aggressive baselines such as NER~\cite{ner} and DP-KSA~\cite{dpksa}, Eraser4RAG preserves substantially more public knowledge and maintains stronger downstream RAG utility, showing a better privacy--utility balance for reusable document-level privacy erasure.
Furthermore, Eraser4RAG has a smaller impact on downstream generation performance than baselines, validating the effectiveness of our Eraser4RAG\footnote{Our code and annotated dataset are available at: \url{https://anonymous.4open.science/r/Eraser4RAG-4E39/}.}.

The innovations of this paper are as follows:
\begin{itemize}
    \item In RAG scenario, we introduce the privacy erasure task, which requires addressing the issues of multi-document de-anonymization, controllable erasure, and available reasoning.
    
    \item We propose Eraser4RAG, the first model for controlled private knowledge erasure for RAG, reducing private knowledge while retaining essential public knowledge through RL.
   
    \item Experimental results show that Eraser4RAG effectively maximizes public triples retention and minimizes private triples, validating the effectiveness of our RL-based strategy.
\end{itemize}

\section{Privacy Erasure Task}

In RAG, the retrieved documents $D = \{ d_i \}_{i=1}^{k}$ may contain private information that needs to be erased before further LLM reasoning. Let $I_{\text{pri}}(D)$ and $I_{\text{pub}}(D)$ denote the private and public information in $D$, respectively. The privacy erasure model is denoted as $\mathcal{M}_{\theta}$, where $\theta$ represents the model parameters. The processed document set is given by $D' = \{ d_i' \}_{i=1}^{k} = \{ \mathcal{M}_{\theta}(d_i) \}_{i=1}^{k}$. Then the objective of privacy erasure is:$$\min_{\theta} \big[|I_{\text{pri}}(D')|,\ -|I_{\text{pub}}(D')|\big].$$ In RAG scenario, the definitions of private and public information, $I_{\text{pri}}$ and $I_{\text{pub}}$, are flexible and random. Due to the challenge of cross-document de-anonymization, it is insufficient to minimize the private information $I_{\text{pri}}(d_i')$ of individual documents alone; instead, the private information of the entire document set must be considered.

We use knowledge graph triples as the representation unit to quantify erasure effectiveness, since it enables: (1) fine-grained control over the removal of private information by precisely locating and manipulating specific knowledge units; and (2) quantitative evaluation of privacy erasure and public information preservation through triple-level comparison. 
For example, “Marie Curie, working with her husband Pierre in a makeshift lab during 1898, was the first to isolate polonium…” can be expressed as: <Marie Curie, collaborated\_with, Pierre Curie>; <Marie Curie, conducted\_research\_\allowbreak in, 1898>; <Polonium, named\_after, Poland>. 
Such fine-grained triples enable targeted deletion and accurate assessment of erasure quality.


\label{define}Both private and public information can be expressed as entity-relation triples $[h, r, t]$, where the head entity $h$ and the tail entity $t$ represent the subject and object of the relation $r$, respectively. In addition, for a non-relational private entity $A$ to be deleted (e.g., only PIIs, temporal dimensions), we represent it in the form of a self-loop triple, that is, $[A, A, A]$. This enables us to uniformly handle both relational and non-relational private knowledge within a triple-based framework. Private and public information within $D$ are thus represented as triple sets $G_{\text{pri}}$ and $G_{\text{pub}}$.
The partitioning of private and public information must satisfy: Neither private information can be inferred from any combination of public information, nor public information from any combination of private information. 
    
    
\begin{figure}[!t]
    \centering
    \includegraphics[width=\linewidth]{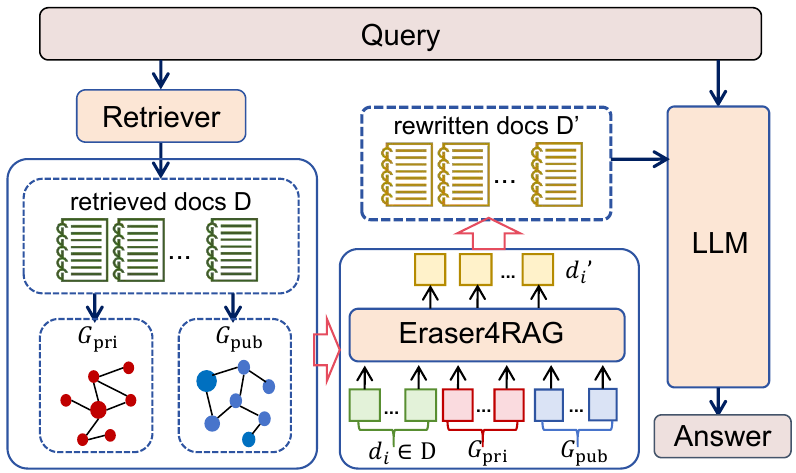}
    \caption{RAG pipeline with privacy erasure.}
    \Description{RAG pipeline with privacy erasure.}
    \label{fig:overview}
\end{figure}

Given this setup, our task is to train a privacy-aware rewriting model $\mathcal{M}_{\theta}$ to transform each document $d_i$ into its rewritten version $d_i'$, ensuring that private information is removed while public information is retained, $d_i' = \mathcal{M}_{\theta}(d_i, G_{\text{pri}}, G_{\text{pub}}).$

For privacy-preserved RAG service,we assume the data controller (e.g., banks or hospitals) defines and annotates private information in accordance with domain-specific regulations. 
The pipeline is shown in Figure~\ref{fig:overview}. Given a query $q$, the retriever first retrieves documents $D = \{ d_i \}_{i=1}^k$. The data controller identifies the private triples $G_{\text{pri}}$ and public triples $G_{\text{pub}}$ from documents. The rewriting model then rewrites each $d_i \in D$ according to $G_{\text{pri}}$ and $G_{\text{pub}}$, producing the rewritten documents $D' = \{ d_i' \}_{i=1}^k$. Given the query $q$ and the rewritten documents $D'$, LLM generates an answer, $a = \text{LLM}(q, D').$

\section{Approach}
\begin{figure}[!t]
    \centering
    \includegraphics[width=\linewidth]{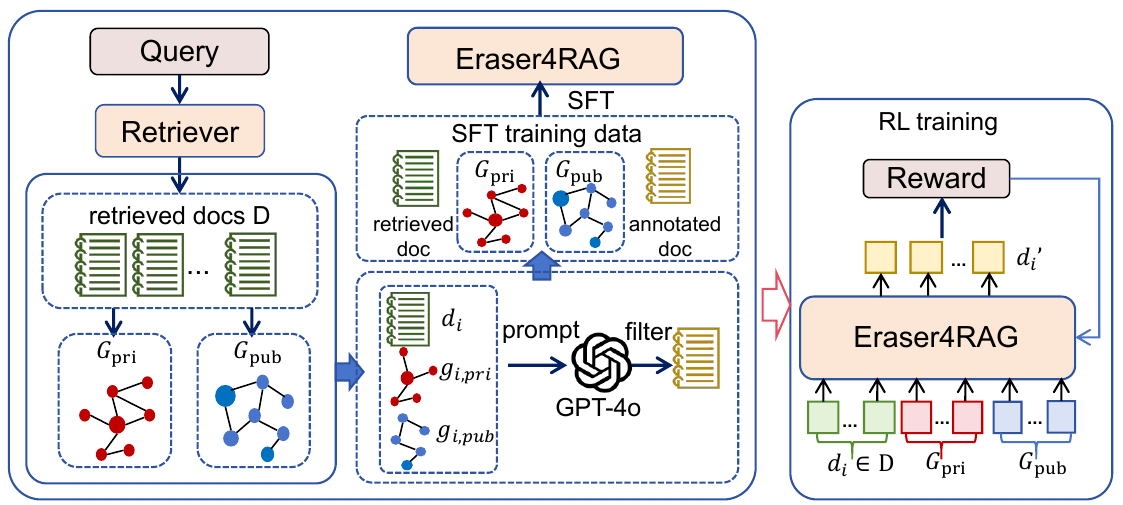}
    \caption{The training process of Eraser4RAG.}
    \Description{The training process of Eraser4RAG.}
    \label{fig:framework}
\end{figure}

The training process of Eraser4RAG consists of two stages: supervised fine-tuning and reinforcement learning (RL), as shown in Figure~\ref{fig:framework}. First, GPT-4o generates rewritten versions of the documents, from which high-quality rewrites are selected. Next, these selected rewrites are used to supervise fine-tuning Flan-T5 model, enabling it to acquire an initial rewriting capability. Finally, the fine-tuned model undergoes multi-objective RL, optimizing it to remove private information while preserving as much public information as possible.
\subsection{Supervised Fine-tuning}
This section outlines the annotated data for training, including privacy definitions, GPT rewrites, and data filtering, followed by the training method.

\subsubsection{Rewritten Data Annotation}
\paragraph{Definition of Private Information} 
To train a model capable of removing private information from documents, it is necessary to define what constitutes private information in the training data. Since the definition of private information is subjective and varies across RAG service providers, we do not designate entity types or relations as private in our training data. Instead, we adopt a random selection strategy to simulate diverse privacy definition scenarios.

Specifically, given a query $q$ and top-$k$ relevant documents $D = \{d_i\}_{i=1}^{k}$, a relation extraction model extracts entity-relation triples $g_i = \{[h, r, t]\}$ from the document $d_i$. These extracted triples from all retrieved documents are merged into a knowledge graph $G$, representing a global set of all triples in documents $D$.
A random subset of triples from $G$ is selected as the candidate private set $G_\text{pri}$\footnote{It is worth noting that the triples appearing in the query are not considered private information.}. The remaining triples in $G$, after excluding those designated as private, form the candidate public information set $G_\text{pub}$. To ensure a clean separation between private and public information(see Section~\ref{define}), we apply the following filtering steps: (1) For $G_\text{pub}$, delete triples whose head and tail entities are connected in the private $G_{\text{pri}}$, since these triples can be inferred with private information. (2) For $G_\text{pri}$, delete triples whose head and tail entities are connected in the public $G_\text{pub}$, since these triples can be inferred from $G_\text{pub}$.

\begin{figure}[t]
    \centering
    \includegraphics[width=\linewidth]{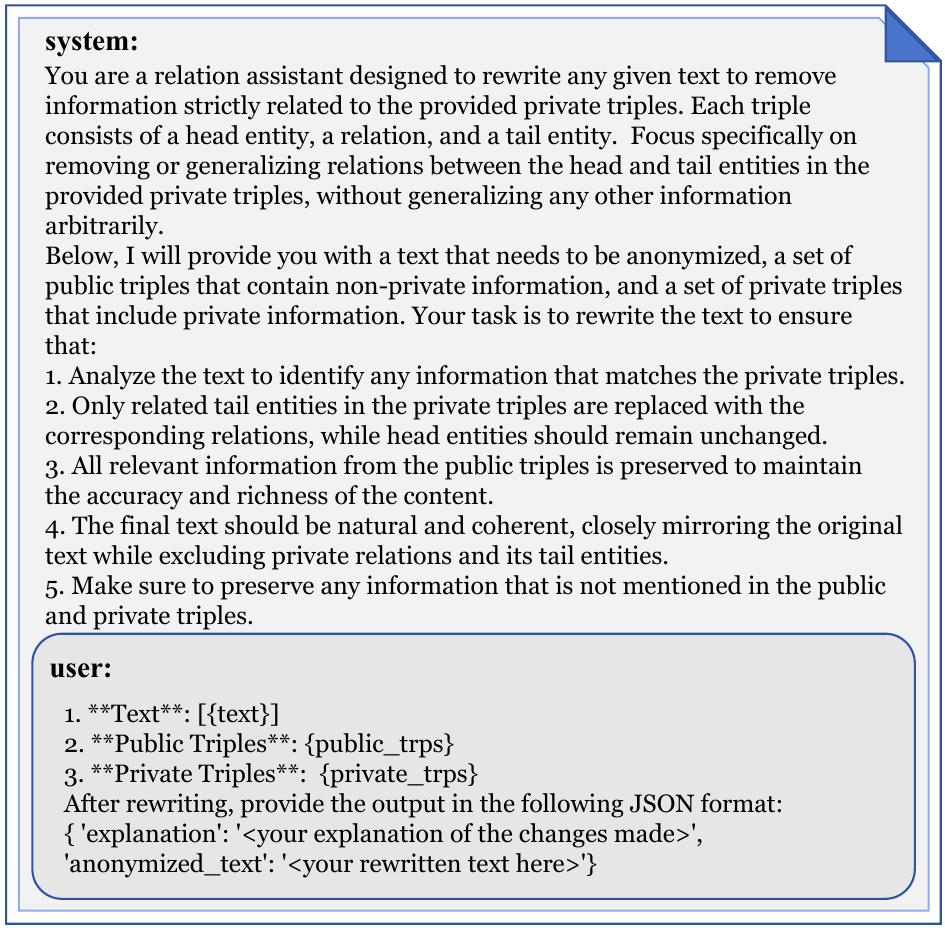}
    \caption{The annotation prompt of GPT-4o.}
    \Description{The annotation prompt of GPT-4o.}
    \label{fig:prompt}
\end{figure}
\paragraph{Rewritten by GPT}
To annotate the training data for the supervised fine-tuning (SFT) stage, we prompt GPT-4o to rewrite a given document based on the provided private and public information sets to remove all private information, retain all public information, and maintain the language style.

Specifically, the private information provided to GPT-4o is the intersection of the entity-relation triples set $g_i$ extracted from document $d_i$ with the global private knowledge set $G_{\text{pri}}$, resulting in $g_{i,\text{pri}} = g_i \cap G_{\text{pri}}$. The public triples are defined similarly as $g_{i,\text{pub}} = g_i \cap G_{\text{pub}}$.
This design reduces the complexity of GPT-4o annotation: instead of requiring GPT-4o to search for document-relevant triples from the entire global graph, the prompt explicitly provides the private and public triples associated with the current document.
The prompt used for data annotation is shown in Figure~\ref{fig:prompt}. Note that the 'private' and 'public' triples are extracted from public datasets, so no privacy is transmitted to API.

\paragraph{Prompt Empirical Tuning}

We experimented with several prompt variants before arriving at the current version, focusing on how well GPT-4o could minimize private triples while preserving the expression of public triples.
In our tuning process, we found the following aspects to be particularly important for optimal performance:

\noindent \textbf{(1) Step-by-step structure:} We explicitly enumerated the rewriting objectives as numbered rules. This significantly improved GPT-4o's consistency and controllability.\\
\textbf{(2) Explanation generation:} Requiring the model to output an explanation of what public triples were preserved served as a form of self-verification. This helped us detect hallucinations or omissions in early drafts and encouraged more faithful preservation of public knowledge.\\
\textbf{(3) Style anchoring:} We emphasized that the rewriting should closely mirror the original style and phrasing of the source document when rewriting public content. Without this, GPT-4o tended to paraphrase unnecessarily, which could affect downstream utility.

To assess the quality of each prompt variant, we measured two key outcomes on a validation set: (1) the proportion of retained private triples in the rewritten text, and (2) the proportion of retained public triples in the rewritten text. The selected prompt yielded the best trade-off between these two, and its effectiveness is reflected in our quantitative results. We have added this point to the revised manuscript.

\paragraph{Data Filtering}
To ensure the quality of the training data, we filter the GPT-4o rewritten documents. Specifically, we apply a relation extraction model to each rewritten document $d_i'$ to extract the set of entity-relation triples $g_i'$. We compute the intersections of $g_i'$ with the original document’s local private set $g_{i,\text{pri}}$ and public set $g_{i,\text{pub}}$, obtaining the private set $g_{i,\text{pri}}'$ and the public set $g_{i,\text{pub}}'$ for $d_i'$.

To quantitatively evaluate the quality of the rewritten documents, we define two retention rates: privacy retention rate $r_{\text{pri}} = \frac{\#g_{i,\text{pri}}'}{\#g_{i,\text{pri}}}$, which measures the proportion of private information retained, and public information retention rate $r_{\text{pub}} = \frac{\#g_{i,\text{pub}}'}{\#g_{i,\text{pub}}}$, proportion of public information retained.
Rewritten documents with $r_{\text{pri}} = 0$ and $r_{\text{pub}} > 0.8$ are selected into the SFT dataset $D_{c}$. 
This filtering criterion ensures that the selected supervision data completely removes local private triples while retaining most public triples, providing a high-quality initialization for subsequent RL optimization.

\subsubsection{Training Rewriting Model}
We fine-tune the pretrained Flan-T5 model on the annotated dataset to equip it with basic privacy removal capabilities. Following a similar approach to \citet{rebel}, we introduce special tokens to linearize the private and public triple sets, ensuring compatibility with the language model’s input format. A public triple $[h_{\text{pub}}, r_{\text{pub}}, t_{\text{pub}}]$ is represented as:$$\texttt{<csubj>} h_{\text{pub}} \texttt{<crel>} r_{\text{pub}} \texttt{<cobj>} t_{\text{pub}} \texttt{<ce>},$$
and a private triple $[h_{\text{pri}}, r_{\text{pri}}, t_{\text{pri}}]$ is represented as:$$\texttt{<rsubj>} h_{\text{pri}} \texttt{<rrel>} r_{\text{pri}} \texttt{<robj>} t_{\text{pri}} \texttt{<re>}.$$
The original document $d$, the linearized global private triples $G_{\text{pri}}$ and public triples $G_{\text{pub}}$ are concatenated to form the input to the rewriter, allowing it to learn to identify all relevant triples related to the document content from the global set, rather than solely processing the privacy information defined within a single document. This design enables the model to handle cross-document de-anonymization challenges effectively.

Given an original document $d$, we denote $d_+ \in D_{c}$ as the rewritten version produced by GPT. The objective of supervised fine-tuning is to optimize the parameters $\theta$ of the rewriter $\mathcal{M}_{\theta}$ by minimizing the cross-entropy loss between the model’s prediction $d'$ and the target $d_+$:
$$ 
L_{sft} = - \sum_{t=1}^T y_{d+}^{(t)} \log y_{d'}^{(t)},
$$ 
where $T$ is the number of tokens, $y_{d+}^{(t)}$ is the one-hot vector for the $t$-th token in the target sequence, and $y_{d'}^{(t)}$ is the predicted token probability distribution.

\subsection{Reinforcement Learning}
Although the rewriter acquires a basic ability to remove relevant private information after supervised fine-tuning, further optimization is required to retain sufficient essential public information while removing as much private information as possible.
We employ RL framework with PPO algorithm \cite{ppo_1,ppo_2} to further optimize Eraser4RAG. 
We formulate rewriting as a sequence-level decision process, where the state consists of the input document and generated prefix, the action is the next token sampled from the rewriter, and the reward is computed after the complete rewritten document is generated.

The reward function $R$ is designed based on the privacy retention rate $r_{\text{pri}}$ and the public information retention rate $r_{\text{pub}}$, evaluating the rewriting:
$$R = r_{\text{pub}} \cdot \exp(-p \cdot r_{\text{pri}}),$$
where $r_{\text{pub}}$, $r_{\text{pri}} \in [0,1]$, and $p$ is a dynamically adjusted parameter. 
$R$ balances two competing objectives: maximizing the retention of public information while minimizing the presence of private information. 
The exponential penalty term $\exp(-p \cdot r_{\text{pri}})$ ensures that even a small amount of retained private information significantly reduces the reward, strongly encouraging the model to eliminate all privacy-sensitive content. At the same time, the linear dependence on $r_{\text{pub}}$ incentivizes the preservation of as much public information as possible. 
The parameter $p$ controls the privacy–utility trade-off.

\section{Experiments}

\subsection{Tasks and Evaluation Metrics}
Regarding the challenges we introduce the experimental tasks and evaluation metrics.

\subsubsection{Split of Private and Public Triples}
In the evaluations of Private Information Erasure, Public Information Preservation, and Resistance to De-Anonymization, the split of private and public triples follows the same random sampling procedure as in the training set. For the downstream RAG, we filter the private triples by removing any triple that explicitly represents a question–answer relationship, ensuring the queries are harmless. For the Resistance to Privacy Attacks evaluation, the private triples are directly constructed from the question–answer pairs.

\subsubsection{Privacy Removal and Public Knowledge Preservation}
To evaluate the erasing performance, we select the specialized test set $D_{\text{special}}$ from the retrieved documents set $D$:
\begin{equation}  
\begin{aligned}  
D_{\text{special}} = &\{ d_i \in D \mid \exists\; [h, r, t]_{\text{pub}} \in g_{i,\text{pub}}, \\  
& [h, r, t]_{\text{pri}}\in g_{i,\text{pri}},  
 \text{s.t. } t_{\text{pub}} = t_{\text{pri}} \}.  
\end{aligned}  
\end{equation} 
$D_{\text{special}}$ consists of documents where public and private information share the same tail entity but different head entity or relation. Experiments on $D_{\text{special}}$ mimic a scenario to remove private relationships while retaining public information, for fine-grained evaluation of anonymization. 

We employ private information retention rate $r_{\text{pri}}$ and public information retention rate $r_{\text{pub}}$ as assessing metrics:
$$r_{\text{pri}} = \#g_{i,\text{pri}}'/{\#g_{i,\text{pri}}};\; r_{\text{pub}} = \#g_{i,\text{pub}}'/{\#g_{i,\text{pub}}},$$ where $\#g_{i,\cdot}'$ is the number of triples remaining in the rewritten document $d_i'$, and $\#g_{i,\cdot}$ is the original count of triples in $d_i$. A lower $r_{\text{pri}}$ indicates better privacy removal, and a higher $r_{\text{pub}}$ signifies better public information preservation. 

In addition, we consider a standalone privacy removal task for entities that cannot be expressed as triples. Anonymization is successful if the entity no longer appears in the rewritten document. We define the entity removal rate as $r_{\text{remove}} = N_{\text{remove}} / N_{\text{total}}$, where $N_{\text{remove}}$ is the number of documents which private entities are successfully removed, and $N_{\text{total}}$ is the total number of documents to be anonymized. Fluency of rewritten texts is measured using perplexity (PPL)~\cite{ppl}, where lower PPL means higher fluency.
\subsubsection{Resistance to De-Anonymization}
To assess whether the erasing method can resist cross-document de-anonymization, we construct a retrieval document subset that is susceptible to such attacks. The inference risk subset is:  
\begin{equation}  
\begin{aligned}
\{D\}_{\text{infer}} &= \{ D \mid \exists \;  d_i \in D,\;(h, t) \in g_{i,\text{pri}}\\
s.t. \;& h\not\sim_{G_{\text{pri}} \setminus g_{i,\text{pri}}}t
\land  h \sim_{G_{\text{unpri}} \setminus g_{i,\text{unpri}}} t \},
\end{aligned}  
\end{equation}
where $G_{\text{unpri}} = G \setminus G_{\text{pri}}$ is global unprivate knowledge graph of documents set $D$, and $g_{i,\text{unpri}} = g_i \setminus g_{i,\text{pri}}$ is unprivate triple set of individual document $d_i$. $G_{\text{unpri}} \setminus g_{i,\text{unpri}}$ represents the unprivate knowledge graph of other documents $D\setminus\{d_i\}$. It simulates cases where privacy is removed from one document but remains inferable through non-private information in other documents.  

We evaluate de-anonymization risk on $\{D\}_{\text{infer}}$ by constructing an entity-relation graph $G'$ based on the anonymized document set $D'$. For each privacy-sensitive triple $[h, r, t]_{\text{pri}} $ in the original private triple set $ G_{\text{pri}}$, if $h_{\text{pri}}$ and $t_{\text{pri}}$ remain connected in $G'$ via intermediate entities and relations from other documents, we consider the private information $[h, r, t]_{\text{pri}}$ to be potentially inferable. Therefore, we measure the risk of de-anonymization using the privacy connection ratio:  
$$r_{\text{connect}} = \# \{ [h, r, t]_{\text{pri}}\in G_{\text{pri}} \mid h \sim_{G'} t \} /{\# G_{\text{pri}}}.$$
This evaluation inherently accounts for multi-hop reasoning. 
A higher $r_{\text{connect}}$ indicates a higher risk due to multi-document reasoning. 
\subsubsection{Utility on Downstream Task}
To evaluate the utility of documents once anonymized, we assess their performance on downstream tasks using four QA datasets. Specifically, we measure the accuracy (acc) when the question and answer do not involve private information, that is, when the entities in the question and answer are not connected in the private knowledge graph $G_{\text{pri}}$.
\subsubsection{Resistance to Privacy Attacks}
Previous work~\cite{rw7} evaluates privacy leakage by prompting LLMs to reproduce private content (e.g., phone numbers, specific sentences). However, our scenario involves multi-document contexts where private knowledge often emerges through entity–relation associations across documents, rather than appearing in a single sentence.
We design a QA-based privacy attack, where question–answer pairs reflect private knowledge. We first select those queries that LLM can answer before anonymization, and construct the privacy triples accordingly. During evaluation, anonymized documents are provided, and the accuracy of these queries defines the attack success rate $r_{\text{attack}}$. Lower $r_{\text{attack}}$ indicates stronger resistance to inference-based privacy attacks.

\subsection{Experimental Settings}
\subsubsection{Datasets}
We conduct model training and evaluation on four QA datasets: PopQA~\cite{popqa}, TriviaQA~\cite{tqa}, Natural Questions (NQ)~\cite{nq}, and HotpotQA~\cite{hqa}, retrieving relevant documents from the Wikipedia corpus. 
Additionally, we validate the model's transferability on the NewsQA~\cite{newsqa} dataset. 
PopQA is a knowledge-intensive QA dataset with questions derived from structured knowledge bases.   
TriviaQA consists of trivia-style questions sourced from online trivia websites, featuring diverse topics.  
NQ contains real-world queries from Google Search.  
HotpotQA is a multi-hop QA dataset requiring reasoning over multiple documents. For the above dataset, use the provided training set to perform GPT-4o annotation and data filtering, and use the test dataset to conduct performance testing of the rewritten model. NewsQA a reading comprehension dataset built from CNN news articles.

\subsubsection{Baselines}
To validate the effectiveness of Eraser4RAG, we compare it with seven baseline methods. The self-disclosure abstraction model (Abstraction)~\cite{rw10} rephrases sensitive self-disclosure spans into less specific terms, generating diverse alternatives that preserve privacy. Anonymizer \cite{rw12} proposes a feedback-guided adversarial anonymization framework to iteratively anonymize text. IncogniText \cite{rw11} rewrites text with incorrect private attribute values to mislead adversaries. NER \cite{ner} identifies and replaces all entities in the document. Additionally, we employ two GPT-4o~\cite{gpt} rewriting strategies: gpt-all using the global private triple set $G_{\text{pri}}$ and public triple set $G_{\text{pub}}$ as prompts, and gpt-each utilizing only the private triple set $g_{i,\text{pri}}$ and public triple set $g_{i,\text{pub}}$ from the individual retrieved document $d_i$. The GPT-4o prompt is provided in Figure~\ref{fig:prompt}. 
We further include DP-KSA~\cite{dpksa}, a differentially private RAG framework that retrieves top-ranked documents, generates an ensemble of responses from the LLM, and privately selects the most frequent keywords via the propose-test-release paradigm. The selected keywords are then augmented into the query to produce the final output. Notably, DP-KSA runs in a query-dependent manner, extracting keywords relevant to each specific query rather than producing a reusable sanitized document. We evaluate it under privacy budgets $\epsilon=1,3,5$ with the same retriever and generator.
Since several baselines are not originally designed for controlled private-public triple rewriting, we adapt them to our setting as follows.
\paragraph{Abstraction} The original Abstraction approach is not designed to operate under a predefined private-public knowledge distinction. To make it applicable as a baseline in our privacy-focused setting, we adapted it by identifying the tail entities from each private triple as the target spans for anonymization. These entity spans were then fed into the anonymization model from Abstraction, which generates substitute spans intended to mask the identity of the original entities. Finally, we applied a regular-expression-based replacement procedure to substitute the entity mentions in the original text with the anonymized spans suggested by the model. This adaptation allowed us to evaluate Abstraction's anonymization capability under a controlled setting where privacy is explicitly defined, enabling a more direct comparison with our method.
\paragraph{Anonymizer} This work introduces an adversarial anonymization framework that leverages LLMs’ strong inferential capabilities to detect and remove private information. Specifically, the method operates in two stages: (1) adversarial inference LLM first identifies potentially sensitive content in a text, and (2) anonymization LLM rewrites or removes such content to reduce the risk of disclosure while maintaining the overall utility of the text. In our experiments, we define privacy using the relation and tail entity to determine what information should be considered private. The anonymization process then aims to eliminate or obscure these private triples from the text, while preserving all other public knowledge.
\paragraph{Incognitext} Incognitext also employs two LLMs: one LLM infers the private attributes from a document, and the second LLM rewrites the document using incorrect private attribute values to mislead the adversarial LLM. The process involves two rounds of interaction to achieve anonymized rewriting of the document’s private attributes. In our adaptation, for each document, we treat the relations in its private triples as the private attributes, the tail entities as the correct private attribute values, and use "unknown" as the incorrect private attribute values during rewriting.
\paragraph{NER} For each document, NER is performed using a BERT-based named entity recognition model to identify all entities in the text, which are then replaced with their corresponding identifiers.

\subsubsection{Implementation Details}
We use Llama3-8b-instruct as the base LM for the generator and Flan-T5-large~\cite{flant5}  as the base model for privacy-preserving text rewriting. Our rewriting model only has 770M parameters, which can run locally in resource-constrained settings. For retrieval, we adopt the off-the-shelf Contriever-MS MARCO model~\cite{contriever} by default, retrieving up to ten documents for each input. The relation extraction model is relik-relation-\allowbreak extraction-small~\cite{relik}. 
PopQA retrieval documents are used for GPT-4o annotation and filtering, and the initial sampling rate for privacy triples is 25\%. For SFT, we employ the AdamW optimizer \cite{adam} and train for 3 epochs with a learning rate of 5e-5. During RL phase, the discount factor $\gamma=0.99$ and the learning rate is 1e-5. The hyperparameter $p$ in the reward function $R$ is initiated as 20 and increases by 5 per 350 iterations to 40. RL training data includes retrieval documents from PopQA, TriviaQA, NQ, and HotpotQA.

\subsection{Entity Variations, Matching Heuristics, and Limitations}
In real-world applications, entities may appear in various surface forms, including abbreviations, nicknames, or indirect references. \textbf{However, these various entities are precisely the challenge that researchers in knowledge extraction field aim to address, and our paper focuses on evaluating a model’s ability to erase private knowledge}. To achieve this, we adapt the state-of-the-art model ReLiK~\cite{relik} from knowledge extraction field to ensure that the extraction tool can reliably and comprehensively identify knowledge from the documents. This model achieves state-of-the-art performance in both in-domain and out-of-domain benchmarks while using academic budget training and with up to 40x inference speed compared to competitors; 

And we use the formal and well-structured Wikipedia corpus for our experiments. Entity references are typically consistent or only moderately abbreviated, e.g., “George James Rankin” may later appear as “George” or “George Rankin,” but not as unrelated nicknames. This consistency allows relation extraction to work reliably, and we also implement a simple entity matching heuristic during triple classification: if all tokens of a short form appear in a longer one (i.e., every token in A is present in A'), we consider them as referring to the same entity. This helps prevent cases like <George James Rankin, works\_at, TechVille> being labeled as private while <George, works\_at, TechVille> remains in the public set, reducing the risk of inadvertently treating a private triple as unrelated to a semantically equivalent public one.

That said, we acknowledge that in more informal or noisy text domains, such as social media or medical records, surface forms of entities may vary significantly. In such scenarios, more sophisticated techniques like coreference resolution, alias detection, or embedding-based entity normalization would be required before relation extraction, to ensure robust privacy detection and erasure.

\subsection{Main Results}
\subsubsection{Private Information Erasure}

Table \ref{tab:entity} compares standalone private entity removal. Eraser4RAG achieves higher $r_{\text{remove}}$ than baselines with better fluency, indicating that the self-loop triple formulation effectively supports the removal of standalone entities. Abstraction method attains 100\% removal rate by direct span replacement, but leads to ungrammatical output and lower fluency. 
DP-KSA shows decreasing $r_{\text{remove}}$ as $\epsilon$ increases, which is counter-intuitive for DP mechanisms. This occurs because DP-KSA operates on the keyword level: the propose-test-release mechanism selects keywords based on their frequency across ensemble responses, and private entities that coincidentally match high-frequency keywords (e.g., common names appearing across documents) may survive the selection process. At higher $\epsilon$, more keywords pass the PTR test, increasing the chance that private entities slip through. Furthermore, since DP-KSA only retains sparse keywords rather than coherent sentences, it yields abnormally high PPL values, reflecting the loss of fluency in the rewritten output.
GPT-4o shows high fluency with 200B parameters but a lower removal rate than Eraser4RAG with 770M parameters, confirming Eraser4RAG’s consistent handling of both relational and standalone privacy.

Table \ref{tab:qa_pri} compares the private information retention rate $r_{\text{pri}}$ across models on $D_{special}$ test set from four QA datasets. 
Among rewriting-based baselines, Eraser4RAG consistently achieves the lowest $r_{\text{pri}}$; NER and DP-KSA obtain even lower values, but through aggressive removal strategies (entity masking or keyword suppression) that severely compromise public information retention.
Among the other baselines, gpt-all exhibits the highest retention rate, suggesting that using a global privacy-aware prompt for GPT-4o is less effective in eliminating sensitive information compared to document-specific rewriting. Abstraction model remains less effective than Eraser4RAG due to replacing private spans is unable to fully obscure sensitive relations.
Compared with Abstraction, gpt-each, and gpt-all, Eraser4RAG yields lower private triple retention on all four datasets, indicating that relation-aware rewriting is more effective than span-level abstraction or prompt-only rewriting.

\begin{table}[t]
\centering
\caption{Performance comparison of different models for private entities in $r_{\text{remove}}\uparrow$ and $\text{PPL}\downarrow$.}
    \begin{tabular}{lccc}
        \toprule
        Model & Abstraction & Anonymizer & IncogniText\\
        \midrule
        $r_{\text{remove}}$(\%) & 100 & 71.6 & 80.4 \\
        PPL & 1.35 &  1.29 & 1.25 \\
        \toprule
        Model & DP-KSA$_{\epsilon=1}$ & DP-KSA$_{\epsilon=3}$ & DP-KSA$_{\epsilon=5}$\\
        \midrule
        $r_{\text{remove}}$(\%) & 74.0 & 70.0 & 66.7 \\
        PPL & 234.06 & 286.57 & 188.48 \\
        \toprule
        Model & NER & GPT-4o & Eraser4RAG\\
        \midrule
        $r_{\text{remove}}$(\%) & 99.8 & 84.7 & 93.7 \\
        PPL & 1.26 & 1.21 & 1.23 \\
        \bottomrule
    \end{tabular}

\label{tab:entity}
\end{table}

\begin{table}[!t]
\centering
\caption{Performance comparison of different models in private and public information retention.}
\begin{subtable}[c]{\linewidth}
    \caption{Comparison of $r_{pri}\downarrow$ across different baselines on $D_{special}$ test set of PopQA, TQA, NQ and HQA.}
    \centering
    \begin{tabular}{lcccc}
        \toprule
        Model & PopQA & TQA & NQ & HQA \\
        \midrule
        Abstraction & 0.1551 & 0.1808 & 0.1489 & 0.1442 \\
        Anonymizer & 0.2463 & 0.2846 & 0.2765 & 0.2593 \\
        IncogniText & 0.2865 & 0.1626 & 0.1522 & 0.1787 \\
        NER & \textbf{0.0180} & \textbf{0.0399} & \textbf{0.0493} & \textbf{0.0176} \\
        gpt-each & 0.1753 & 0.1613 & 0.1884 & 0.1696 \\
        gpt-all & 0.3069 & 0.2467 & 0.2730 & 0.2767 \\
        DP-KSA$_{\epsilon=1}$ & 0.0587  &  0.0608  &  0.0621  &  0.0427 \\
        DP-KSA$_{\epsilon=3}$ & 0.0720  &  0.0684  &  0.0803  &  0.0518 \\
        DP-KSA$_{\epsilon=5}$ & 0.0758  &  0.0751  &  0.0924  &  0.0582 \\
        Eraser4RAG & 0.1099 & 0.1463 & 0.1429 & 0.1436 \\
        \bottomrule
    \end{tabular}

    \label{tab:qa_pri}
\end{subtable}
\begin{subtable}[c]{\linewidth}
    \centering
    \caption{Comparison of $r_{pub}\uparrow$ across different baselines on $D_{special}$ test set of PopQA, TQA, NQ and HQA.}
    \begin{tabular}{lcccc}
        \toprule
        Model & PopQA & TQA & NQ & HQA \\
        \midrule
        Abstraction & 0.4384 & 0.4247 & 0.3997 & 0.4107 \\
        Anonymizer & 0.2386 & 0.2819 & 0.2684 & 0.2477 \\
        IncogniText & 0.4553 & 0.3140 & 0.2927 & 0.3477 \\
        NER & 0.0193 & 0.0427 & 0.0471 & 0.0193 \\
        gpt-each & \textbf{0.4825} & 0.4337 & 0.4512 & 0.4628 \\
        gpt-all & 0.4040 & 0.3323 & 0.3632 & 0.3790 \\
        DP-KSA$_{\epsilon=1}$ & 0.0590  &  0.0598  &  0.0561 &   0.0444 \\
        DP-KSA$_{\epsilon=3}$ & 0.0674  &  0.0668  &  0.0710  &  0.0583 \\
        DP-KSA$_{\epsilon=5}$ & 0.0726  &  0.0715  &  0.0777  &  0.0613 \\
        Eraser4RAG & 0.4793 & \textbf{0.4441} & \textbf{0.4663} & \textbf{0.4720} \\
        \bottomrule
    \end{tabular}

    \label{tab:qa_pub}
\end{subtable}

\end{table}

\subsubsection{Public Information Preservation}

Table~\ref{tab:qa_pub} compares $r_{\text{pub}}$ across models. Eraser4RAG achieves the highest performance on three out of the four datasets (TQA, NQ, and HQA), while gpt-each is slightly better on PopQA. Compared with GPT-based rewriting and anonymization baselines, Eraser4RAG more consistently preserves public relational knowledge, suggesting that explicitly modeling both private and public triples helps avoid unnecessary deletion.

NER and DP-KSA obtain much lower $r_{\text{pub}}$ values. However, this does not indicate better privacy-aware rewriting. NER masks nearly all entities, thereby destroying both private and public triples. DP-KSA operates under a different paradigm: it does not rewrite documents, but privately releases a sparse set of query-relevant keywords. As a result, most document-level relational structure is discarded, making it difficult for the relation extraction model to recover public triples from its output. Therefore, although DP-KSA can reduce private triple exposure, it does so at the cost of severe public knowledge loss.

Overall, these results show that Eraser4RAG achieves a better privacy--utility balance than existing anonymization and GPT-based rewriting baselines: it removes sensitive relational knowledge while preserving public triples needed for downstream generation.

\begin{table}[t]
\centering
\caption{Performance comparison of different models in $r_{\text{connect}}\downarrow$ on $\{D\}_{\text{infer}}$ test set.}
    \begin{tabular}{lcccc}
        \toprule
        Model & PopQA & TQA & NQ & HQA \\
        \midrule
        Abstraction & 0.3728 & 0.4069 & 0.4059 & 0.4135 \\
        Anonymizer & 0.2996 &  0.3024 & 0.4411 & 0.3698 \\
        IncogniText & 0.3923 & 0.4002 & 0.3959 & 0.4284 \\
        NER & \textbf{0.0221} & \textbf{0.0293} & \textbf{0.0341} & \textbf{0.0291} \\
        gpt-each & 0.3665 & 0.3811 & 0.3749 & 0.3775 \\
        gpt-all & 0.4352 & 0.4413 & 0.4366 & 0.4407 \\
        DP-KSA$_{\epsilon=1}$ & \underline{0.0654}  &  \underline{0.0721} &   \underline{0.0681} &   \underline{0.0544}\\
        DP-KSA$_{\epsilon=3}$ & 0.0794  &  0.0831  &  0.0899  &  0.0631\\
        DP-KSA$_{\epsilon=5}$ & 0.0830  &  0.0876  &  0.1005  &  0.0706\\
        Eraser4RAG & 0.2600 & 0.3316 & 0.2102 & 0.2821 \\
        \bottomrule
    \end{tabular}

\label{tab:qa_connect}
\end{table}

\subsubsection{Resistance to De-Anonymization}
Table \ref{tab:qa_connect} compares $r_{\text{connect}}$ across different methods on the corresponding test set $D_{\text{connect}}$ of four datasets.
Except for NER and DP-KSA, Eraser4RAG consistently achieves the lowest $r_{\text{connect}}$ across datasets, 
demonstrating effective prevention of private entity pairs from being reconnected in the anonymized graph. In contrast, gpt-all exhibits the highest $r_{\text{connect}}$, suggesting that aggregating all private and public triples for rewriting results in excessive information interference. 
DP-KSA also achieves a low $r_{\text{connect}}$, especially under smaller privacy budgets. This result is consistent with its keyword-selection mechanism: since the final output only uses a small set of privately released keywords rather than the full retrieved documents, many entity-relation paths are removed before graph reconstruction. However, this low connection ratio should be interpreted together with $r_{\text{pub}}$ in Table~\ref{tab:qa_pub}. DP-KSA breaks private connections mainly by discarding most relational structure, including public knowledge, whereas Eraser4RAG reduces cross-document linkability while preserving substantially more public triples. Thus, Eraser4RAG provides a more practical defense for RAG settings where sanitized documents must remain useful for downstream reasoning.
This suggests that simply providing all global triples to a general-purpose LLM may introduce information interference, making it difficult to distinguish relations that should be removed from those that should be preserved.
Abstraction model shows comparable performance to gpt-each, suggesting that replacing sensitive spans with abstracted terms does not fully break cross-document linkability.  

\subsubsection{Downstream RAG Accuracy}
\begin{table}[t]
\centering
\caption{RAG accuracy$\uparrow$ comparison of different privacy-preserving processing methods across four datasets.}
    \begin{tabular}{lcccc}
        \toprule
        Model & PopQA & TQA & NQ & HQA \\
        \midrule
        w/o rewrite & 0.4292 & 0.6655 & 0.3865 & 0.2690 \\
        Abstraction & 0.3785 & \underline{0.6550} & \textbf{0.3635} & 0.2576 \\
        Anonymizer & 0.2252 &  0.5655 & 0.2179 & 0.2029 \\
        IncogniText & 0.3824 & 0.5986 & 0.3423 & 0.2218 \\
        NER & 0.1992 & 0.1438 & 0.1988 & 0.1284 \\
        gpt-each & \textbf{0.3919} & 0.6484 & 0.3460 & \underline{0.2613} \\
        gpt-all & 0.3868 & 0.6443 & 0.3456 & \textbf{0.2640} \\
        DP-KSA$_{\epsilon=1}$ & 0.3003  &  0.5595  &  0.1948  &  0.2051\\
        DP-KSA$_{\epsilon=3}$ & 0.3199  &  0.6006  &  0.2106  &  0.2055\\
        DP-KSA$_{\epsilon=5}$ & 0.3216  &  0.6128  &  0.2251   & 0.2236\\
        Eraser4RAG & \underline{0.3870} & \textbf{0.6635} & \underline{0.3568} & 0.2585 \\
        \bottomrule
    \end{tabular}

\label{tab:rag}
\end{table}
Table \ref{tab:rag} compares RAG accuracy using rewritten documents. Eraser4RAG maintains competitive performance despite eliminating private relationships. 
For example, on PopQA, Eraser4RAG achieves accuracy close to gpt-each while substantially reducing private triple retention, suggesting that the improved privacy protection does not come at the cost of severe utility degradation. For all rewriting methods, introducing privacy-preserving rewriting generally leads to a slight accuracy drop because some textual evidence is modified or removed.
DP-KSA exhibits a clear privacy--utility trend with respect to $\epsilon$: larger privacy budgets improve downstream accuracy, since more keywords can be released and used in the final prompt. Nevertheless, all DP-KSA variants underperform Eraser4RAG on the four datasets. This is because DP-KSA compresses the retrieved evidence into a small keyword set, which is often sufficient for short-answer QA but less reliable when the answer requires richer context, entity disambiguation, or multi-hop reasoning. In contrast, Eraser4RAG preserves coherent rewritten documents, allowing the generator to access more complete public evidence.
While some answer-relevant connections may be lost due to privacy removal, the model still preserves enough useful information to maintain RAG effectiveness. On the multi-hop dataset HQA, where answers rely on cross-document reasoning, Eraser4RAG’s rewriting tends to remove relations that are crucial for deriving the final answer, leading to a larger performance drop. 
\subsubsection{Resistance to Privacy Attacks}
\begin{table}[t!]
\centering
\caption{Evaluation of privacy leakage under QA-based attack setting.}

    \begin{tabular}{lccccccc}
        \toprule
        Model & Abstraction & Anonymizer & IncogniText \\
        \midrule
        $r_{\text{attack}}\downarrow$ & 0.4001 & 0.3788 & 0.4375 \\
        
        \toprule
        Model & DP-KSA$_{\epsilon=1}$ & DP-KSA$_{\epsilon=3}$ & DP-KSA$_{\epsilon=5}$\\
        \midrule
        $r_{\text{attack}}\downarrow$ & 0.2888 & 0.2974 &0.3137 \\ 
        \toprule
        Model & gpt-each & gpt-all & Eraser4RAG\\
        \midrule
        $r_{\text{attack}}\downarrow$ & 0.4338 & 0.4252 &0.3745 \\
        \bottomrule
    \end{tabular}
\label{tab:attack}
\end{table}
Table \ref{tab:attack} compares privacy leakage in the QA-based attack setting. From the results, Eraser4RAG obtains lower $r_{\text{attack}}$ than most baselines, indicating the effectiveness of privacy protection. 
DP-KSA achieves a lower attack success rate than Eraser4RAG, but its aggressive filtering also renders the documents unusable for legitimate queries, as evidenced by the corresponding drop in RAG accuracy.
Compared with other methods, Eraser4RAG reduces the leakage level by using a privacy erasure approach that targets the relationships between entities, indicating its strong resistance to defend against inference-based privacy attacks.

\subsection{Case Study}
\begin{table}[t]
\caption{The case study comparing with Abstraction baseline.}
\centering
\resizebox{\columnwidth}{!}{
\begin{scriptsize}
\begin{tabular}{@{}p{\linewidth}@{}}
\toprule
\# Case 1:\\
\midrule
\textbf{Document A} :\\
Captain James Cook was a British explorer, navigator, cartographer, and captain in the Royal Navy .... James Cook joined the British merchant navy as a teenager and joined the Royal Navy in 1755. James Cook saw action in the Seven Years' War and subsequently surveyed and mapped much of the entrance... \\
\textbf{Private}: [...<James Cook, participate, Seven Years ' War>...]\\ 
\textbf{Public}: [<James Cook, country, British>, <James Cook, part of, Royal Navy>...]
\\
\midrule
\textbf{Eraser4RAG}: \\
Captain James Cook was a \underline{British} explorer, navigator, cartographer, and captain in the Royal Navy... James Cook joined the \underline{British} merchant navy as a teenager and joined the Royal Navy in 1755. James Cook \textbf{saw action} and subsequently surveyed and mapped much of the entrance...\\
\textbf{Abstraction}: \\
Captain James Cook was a \underline{British} explorer, navigator, cartographer, and captain in the Royal Navy... James Cook joined the \underline{British} merchant navy as a teenager and joined the Royal Navy in 1755. James Cook \textbf{saw action in the conflicting} \textbf{event} and subsequently surveyed and mapped much of the entrance...\\
\midrule
\textbf{Document B} :\\
...He saw action in the Seven Years' War, which was a significant conflict involving the British military. During this time, his contributions to the British forces were well recognized...\\
\textbf{Private}: [...<Seven Years' War, country, British>...]\\
\textbf{Public}: [... <James Cook, country, British>...]
\\
\midrule
\textbf{Eraser4RAG}: \\
... He saw action in \textbf{a significant conflict}. During this time, his contributions to the \underline{British} forces were well recognized...\\
\textbf{Abstraction}: \\
... He saw action in the \textbf{Seven Years' War}, which was a significant conflict involving a \textbf{famous country} military. During this time, his contributions to \underline{the country} forces were well recognized...\\
\bottomrule
\end{tabular}
\end{scriptsize}
}

\label{case-table}
\end{table}

In this case study shown in Table \ref{case-table}, we focus on the private relation: <James Cook, participate, Seven Years' War>,
which is marked as private in Document A. 
In Document A, we successfully removes explicit mentions of James Cook's participation in the Seven Years' War in the rewritten Document A. The original phrase "James Cook saw action in the Seven Years' War" is omitted in the rewritten text, replaced with a more general statement such as "James Cook saw action and subsequently...," thus hiding the private relation. At the same time, public information like James Cook’s nationality and affiliation with Royal Navy is preserved.
In Document B, where <Seven Years' War, country, British> is private, the rewritten text changes the explicit mention "the Seven Years' War" to a vaguer "a significant conflict," aligning with the redacted portions of Document A, thereby concealing the private information while retaining the context of military involvement and contributions to British forces.

However, <James Cook, participate, Seven Years' War> is not explicitly defined as privacy in Doc B. For Abstraction, since privacy is considered only within individual documents rather than the global private triples, the result would include “Seven Years' War”, without hiding <James Cook, participate, Seven Years' War>. Since Abstraction does not perform anonymization based on relationships, and British appears in both public and private relationships, Abstraction replaces British which should be treated as public information.

This case demonstrates that our method can effectively erase targeted private relations from documents while preserving as much public information as possible. It also highlights the importance of addressing privacy from a global, multi-document perspective, as private information might still be inferred if only single documents are considered independently.

\subsection{Ablation Study}
\begin{figure}[t!]
    \centering
    \begin{subfigure}{0.32\linewidth}
        \includegraphics[width=0.99\linewidth]{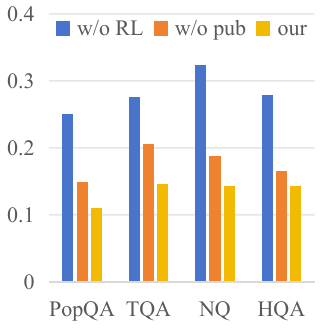}
        \caption{$r_{pri}\downarrow$}
        \label{fig:ablation1}
    \end{subfigure}
    \begin{subfigure}{0.32\linewidth}
        \includegraphics[width=0.99\linewidth]{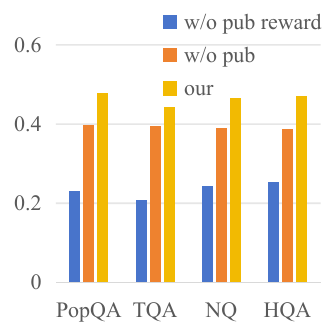}
        \caption{$r_{pub}\uparrow$}
        \label{fig:ablation2}
    \end{subfigure}
    \begin{subfigure}{0.32\linewidth}
        \includegraphics[width=0.99\linewidth]{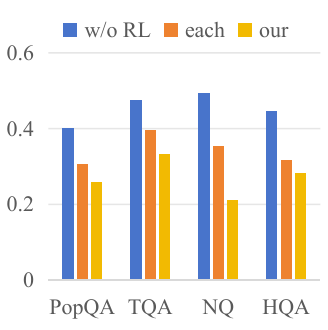}
        \caption{$r_{connect}\downarrow$}
        \label{fig:ablation3}
    \end{subfigure}
    \caption{Ablation study results on different datasets for metrics: (a) $r_{\text{pri}}$, (b) $r_{\text{pub}}$, and (c) $r_{\text{connect}}$. The comparisons include variants without RL (w/o RL), without public triplets (w/o pub), without reward related to public information (w/o pub reward), only inputting triplets from the individual document (each), and our full method (our). }
    \Description{Ablation study results on different datasets for metrics: (a) $r_{\text{pri}}$, (b) $r_{\text{pub}}$, and (c) $r_{\text{connect}}$. The comparisons include variants without RL (w/o RL), without public triplets (w/o pub), without reward related to public information (w/o pub reward), only inputting triplets from the individual document (each), and our full method (our). }
    \label{fig:ablation}
    
\end{figure}
\subsubsection{Public Information Constraint}
To verify the effectiveness of public triples during rewriting, we evaluate when only the original document and the private triple set are provided as input (w/o pub). The results, shown in Figures \ref{fig:ablation1} and \ref{fig:ablation2}, demonstrate the impact on $r_{\text{pri}}$ and $r_{\text{pub}}$. Due to the disjoint nature of public and private information, incorporating public triples when rewriting not only constrains the rewriting from arbitrarily removing content unrelated to private information, but also provides a comparative signal that helps the rewriter better identify privacy.
\subsubsection{RL Training}
To verify the effectiveness of RL, we evaluate the privacy anonymization capability of the rewriter trained only with SFT, as shown in Figure \ref{fig:ablation1} and \ref{fig:ablation3}. Due to limitations in the quantity and quality of annotated data used for SFT, a rewriter trained solely with SFT fails to effectively remove privacy-sensitive relations and prevent cross-document de-anonymization. This further highlights that RL can guide the model toward the desired objective through reward feedback.
\subsubsection{Multi-objective Reward}
To verify the necessity of multi-objective rewards in RL, we evaluate the rewriter’s ability when using only the privacy-related reward $R = \exp(-p \cdot r_{\text{pri}})$, as shown in Figure \ref{fig:ablation2}. It can be observed that omitting the constraint on $r_{\text{pub}}$ in the reward function leads to excessive removal of non-private content, reducing the model’s ability to preserve relevant public information and potentially impacting downstream task performance.
\subsubsection{Global Triple Set}
To validate that using the global triple sets $G_{\text{pri}}$ and $G_{\text{pub}}$ helps mitigate cross-document de-anonymization, we evaluate the performance when prompting the rewriter with only the triples $g_{i,pri}$ and $g_{i,pub}$ relevant to the single document $d_i$, as shown in Figure~\ref{fig:ablation3}. The results indicate that relying solely on document-specific triples significantly increases the risk of cross-document de-anonymization. Therefore, providing the global triple set $G_{\text{pri}}$ imposes stronger constraints on the rewriter, ensuring the elimination of relations that could infer private information beyond those explicitly present within the given document.

\subsection{Transferability}
\begin{table}[t]
\centering
\caption{Erase performance on NewsQA.}
    \begin{tabular}{lccc}
        \toprule
        model & $r_{\text{pub}}\uparrow$ & $r_{\text{pri}}\downarrow$ & $r_{\text{connect}}\downarrow$ \\
        \midrule
        Abstraction &  0.3332 & 0.1075 & 0.4345 \\
        gpt-each &  \underline{0.4272} & 0.1636 & 0.4150 \\
        gpt-all &  0.3321 & 0.2558 & 0.4584 \\
        DP-KSA$_{\epsilon=1}$ & 0.0219  &  \underline{0.0266}   &   \underline{0.0437}\\
        DP-KSA$_{\epsilon=3}$ & 0.0269   & \textbf{0.0244}  &    \textbf{0.0362}\\
        DP-KSA$_{\epsilon=5}$ & 0.0268  &  0.0288   &   0.0471\\
        Eraser4RAG & \textbf{0.4355} & 0.1124 & 0.2622 \\
        \bottomrule
    \end{tabular}

\label{tab:newsqa}
\end{table}
To verify the transferability of Eraser4RAG trained on the Wikipedia corpus, we evaluate its performance on NewsQA dataset, which consists of CNN news articles. 

The results in Table~\ref{tab:newsqa} show that Eraser4RAG achieves the highest public information retention rate, demonstrating that the learned rewriting strategy can generalize beyond Wikipedia-style documents. Although DP-KSA obtains lower $r_{\text{pri}}$ and $r_{\text{connect}}$, its $r_{\text{pub}}$ is close to zero, indicating that the reduction in privacy leakage mainly comes from discarding most document-level relational content rather than selectively preserving useful public knowledge. Therefore, DP-KSA is effective as a query-time private keyword release mechanism, but it is less suitable for our setting where the goal is to produce reusable sanitized documents for RAG.
Compared with Abstraction and GPT-based rewriting baselines, Eraser4RAG better preserves public triples while maintaining competitive privacy removal, showing stronger transferability for document-level privacy erasure.

\section{Privacy Processing Overhead}
\begin{table}[t]
\centering
\caption{Inference time of privacy-preserving processing per query/document set (s).} 
    \begin{tabular}{ccccc}
        \toprule
        Abstraction & gpt-each & gpt-all & DP-KSA & Eraser4RAG  \\ 
        \midrule
        2.2633 & 1.7856 & 1.8659 & 7.2245 & 0.8024\\
        \bottomrule
    \end{tabular}

\label{tab:time}
\end{table}
We report the computational overhead introduced by the privacy-preserving processing stage.
DP-KSA incurs substantially higher inference time (7.22s) because it performs query-time private keyword selection. It first generates an ensemble of $N$ LLM responses, one for each retrieved document, and then constructs a token histogram for PTR-based keyword release. This process must be repeated for every incoming query. In contrast, Eraser4RAG produces static rewritten documents that can be reused across multiple queries once the private knowledge definition is fixed.
Since Eraser4RAG is based on the lightweight Flan-T5-large model, its inference time is significantly lower. This efficiency makes it more practical for real-world applications where computational cost is a concern. Moreover, the reduced inference time does not come at the expense of privacy preservation, as Eraser4RAG maintains strong privacy filtering capabilities while ensuring stable performance in downstream tasks.

\section{Related Work}
Existing approaches to privacy protection primarily include methods targeting model memorization leakage and those addressing retrieval-phase privacy risks. 
To mitigate model memorization leakage, differential privacy~\cite{rw13,rw14} introduces carefully calibrated noise to training data, reducing the risk that sensitive information is memorized. For example, \citet{rw13} fine-tunes models using task-aligned objectives, carefully tuned hyperparameters, and a memory-efficient gradient clipping technique called Ghost Clipping. \citet{rw14} combines DP-SentencePiece, which adds noise to word counts for private tokenization, with DP-Training, which applies gradient clipping and noise during training. Another approaches involve PII identification and anonymization techniques \cite{rw8,rw9} preprocess training data by detecting and removing personally identifiable information. \citet{rw8,rw9} use automated tools (Presidio \cite{Presidio}, Spacy \cite{spacy}) and regular expression-based approaches to detect high-risk personal information in training corpora.

To address inference-phase privacy risks, synthetic data generation \cite{rw15,rw16} replaces sensitive content with artificial data that maintains the statistical properties of real information. \citet{rw15} proposes a two-stage framework for generating synthetic data. The first stage uses attribute extraction and generation to preserve key information from the original data throughout the database, while the second stage employs agent-based iterative refinement to enhance privacy. \citet{rw16} proposes to generate differentially private synthetic few-shot examples from private datasets. Data sanitization and anonymization \cite{rw10,rw11,rw12} preprocess documents by removing or obfuscating sensitive information before they are fed into the model. \citet{rw10} rephrases sensitive self-disclosure spans into less specific terms using fine-tuned language models to generate diverse alternatives. \citet{rw12} proposes a feedback-guided adversarial anonymization framework that leverages the inferential capabilities of LLMs to iteratively anonymize text. Privacy-aware retrieval mechanisms \cite{rw5,rw17} further modify the retrieval process to prevent privacy exposure while preserving enough public information. \citet{rw5} decouples key and query encoders, isolating sensitive data exposure. \citet{rw17} restricts the use of private documents in subsequent public retrievals and ensures queries relying solely on private information remain local. 

However, existing privacy protection methods suffer from key limitations when applied to RAG. 
Naively applying DP to RAG may introduce substantial noise and degrade utility, especially when precise knowledge retention is required. Recent query-time DP-RAG methods such as DP-KSA~\cite{dpksa} mitigate this issue by reducing the generation space to a private keyword-selection problem. However, DP-KSA releases only query-dependent keywords rather than sanitized documents, and therefore does not directly address fine-grained, user-defined private-public knowledge erasure or reusable document-level sanitization.
PII anonymization, lacking dynamic privacy control, cannot adapt to scenario-specific definitions of privacy. Additionally, synthetic data generation requires modifying the retrieval corpus in advance, making database maintenance costly while potentially harming retrieval accuracy due to distorted semantic relationships. Other methods like data sanitization and privacy-aware retrieval operate on isolated documents or queries, overlooking the risk of multi-document de-anonymization. Therefore, we propose Eraser4RAG to perform privacy erasure on retrieved documents. We utilize knowledge graph triples to represent information, which enables precise control over privacy removal while minimizing unnecessary knowledge loss. Our approach leverages the global privacy knowledge graph to effectively prevent multi-document de-anonymization risks.  

\section{Conclusion}
This paper introduces the Privacy Erasure Task and proposes a novel model, Eraser4RAG, for controlled private knowledge erasure in RAG. We address key challenges including multi-document de-anonymization, controllable erasure, and maintaining downstream task utility. Our approach leverages triples to represent information and ensure fine-grained control over privacy erasure while minimizing unnecessary information loss. We further propose RL to effectively remove private triples while retaining essential public triples. 
In future work, we plan to avoid LLMs generating private information, potentially leveraging unlearning to differentiate private content during generation.
In future work, we will explore the use of contextual embeddings combined with adaptive attention strategies to detect and eliminate sensitive material from unprocessed textual data automatically.



\bibliographystyle{ACM-Reference-Format}
\bibliography{sample-base}

\appendix

\end{document}